\documentclass[twoside]{article}
\usepackage[accepted]{aistats2021}

\usepackage{tikz}
\usetikzlibrary{matrix}
\usepackage{algorithm}
\usepackage{algorithmic}
\usepackage[utf8]{inputenc} 
\usepackage[T1]{fontenc}    
\usepackage{hyperref}       
\usepackage{url}            
\usepackage{booktabs}       
\usepackage{amsfonts}       
\usepackage{nicefrac}       
\usepackage{microtype}      

\usepackage{multirow}

\usepackage{environ}
\usepackage{graphicx}
\usepackage{grffile}

\usepackage{caption}
\usepackage{subcaption}
\usepackage{amsmath}

\usepackage{tikz} 
\usepackage{tikzscale}
\usepackage{mathtools}
\usepackage{natbib}
\usepackage{bbm}
\usepackage{array}

\usepackage[export]{adjustbox}

\newcommand{\given} { \,|\, }

\newcommand{\tr}{^{\text{T}} }

\newcommand{\bs}[1]{\boldsymbol{#1}}

\newcommand{\x}{\mathbf{x}}
\newcommand{\z}{\mathbf{z}}
\newcommand{\f}{\mathbf{f}}
\newcommand{\y}{\mathbf{y}}
\newcommand{\iu}{\mathbf{u}}

\newcommand{\am}{\mathbf{m}}
\newcommand{\aL}{\mathbf{L}}
\newcommand{\X}{\mathbf{X}}
\newcommand{\Z}{\mathbf{Z}}

\newcommand{\Y}{\mathbf{Y}}

\newcommand{\id}{\text{d}}

\newcommand{\Normal}[1] { \mathrm{N} {\left(#1\right)}  }

\DeclareMathOperator*{\argmin}{arg\,min}
\DeclareMathOperator*{\argmax}{arg\,max}
\DeclareMathOperator{\E}{\mathbb{E}}

\usepackage{xspace}
\newcommand{\acr}[1]{\textsc{#1}\xspace}

\newcommand{\gp}{\acr{gp}}

\newcommand{\svgp}{\acr{svgp}}

\usepackage{hyperref}
\hypersetup{breaklinks=true,
            colorlinks=true,
            linkcolor=black,
            citecolor=black,
            filecolor=black,
            urlcolor=black}

\author{
Eero Siivola\\
\and Javier Gonz{\'a}lez\\
\and Andrei Paleyes \\
\and Aki Vehtari
}

\begin{document}
\twocolumn[

\aistatstitle{Good practices for Bayesian Optimization of high dimensional structured spaces}

\aistatsauthor{ Eero Siivola \And Javier Gonz{\'a}lez \And Andrei Paleyes \And Aki Vehtari }

\aistatsaddress{ Aalto University \And Microsoft \And University of Cambridge \And Aalto University } ]
\begin{abstract}
The increasing availability of structured but high dimensional data has opened new opportunities for optimization. One emerging and promising avenue is the exploration of unsupervised methods for projecting structured high dimensional data into low dimensional continuous representations, simplifying the optimization problem and enabling the application of traditional optimization methods. However, this line of research has been purely methodological with little connection to the needs of practitioners so far. In this paper, we study the effect of different search space design choices for performing Bayesian Optimization in high dimensional structured datasets. In particular, we analyse the influence of the dimensionality of the latent space, the role of the acquisition function and evaluate new methods to automatically define the optimization bounds in the latent space.  Finally, based on experimental results using synthetic and real datasets,  we provide recommendations for the practitioners.
\end{abstract}

\section{Introduction} \label{sec:introduction}
Let $\mathcal{X}$ be an input space and $f: \mathcal{X} \to \mathbb{R}$ be a continuous \emph{black-box} function. We are interested in solving the global optimization problem of finding the unknown minimum of $f$:
\begin{equation}\label{eq:problem}
\x_{\text{min}} = \argmin_{\x \in \mathcal{X}} f(\x) 
\end{equation}
We make three assumptions:
\begin{enumerate}
\item We can query $f$ using noisy queries $ y = f(\x) + \epsilon$, where $\epsilon \sim \Normal{0, \sigma}$.
\item $\mathcal{X}$ is structured and  high dimensional.
\item  We can access a large unlabelled dataset $\mathbf{X}$ in the input space s.t. $X_i \in \mathcal{X} \: \forall \: i = 1 \ldots N$. 
\end{enumerate}
The goal is to find $\x_{min}$ by limiting the number of queries, which are assumed to be expensive.

These kinds of problems are ubiquitous, as many problems in bioinformatics, chemical engineering and computer science involve optimizing structured objects, such as graphs or images. One important example is designing new molecules, which can be laborious since testing the chemical properties requires wet room experiments, that require manual work by an expert and expensive special equipment.

In traditional applications with a low number of continuous parameters, Bayesian Optimization (BO) is the de facto solution for these gradient-free black-box optimization problems. The core idea of BO is to build a surrogate probabilistic model that efficiently guides the sequential acquisition of new data. However, building and optimizing these probabilistic surrogates in structure high dimensional spaces is challenging and often leads to poor performance.

The recent increase in the availability of high volume datasets has made it possible to use semi-supervised deep generative models to efficiently embed structured high dimensional objects into a lower-dimensional Euclidean space. These models have enabled the use of gradient-free optimization methods for optimizing the structure in the low dimensional manifold (\cite{griffiths2017constrained,kusner2017grammar}). However, this research has so far been merely methodological and with no emphasis on how to design the optimization task itself. There is no research on a) how to decide on the dimensionality of the low dimensional embedding and how this decision affects the optimization task; b) how to optimize the acquisition function in the low dimensional manifold; and c) how to balance between exploration and exploitation when selecting new points. We systematically study the effects of these design choices applied to a variety of high dimensional structured optimization tasks. We hope that our findings will help practitioners to better design their high dimensional structured optimization problems.

The rest of the paper has the following structure. Section \ref{sec:related_work} presents the related work and section \ref{sec:theory} introduces the theoretical background. Section \ref{sec:bovae} describes the framework to perform BO by exploiting deep generative models and the main design choices analysed in the paper.   Section \ref{sec:results}, describes the experimental set-up and presents the main results. Finally, in section \ref{sec:conclusion}, the paper is concluded with discussion and recommendations for the practitioners.

\section{Related work}\label{sec:related_work}

Bayesian Optimization (BO) is considered as the method of choice for sample-efficient gradient-free optimization of low dimensional Euclidean spaces \cite{brochu2010tutorial,shahriari2015taking}. The biggest limitation of BO has been its scalability with respect to the dimensionality of the search space.

The earliest solutions to scale BO to higher-dimensional spaces are based on projecting the input space to low dimensional spaces using linear transformations. \cite{wang2013bayesian} use random linear projections where the BO is performed. \cite{garnett2013active} optimize the linear projection during the optimization to further improve the performance.  The main disadvantages of these methods are that they are only able to find linear manifolds

 Another strategy for high dimensional optimization is to better understand the structure of the search space and impose additional assumptions that simplify the problem. \cite{kandasamy2015high} assume that the search space is composed of disjoint low dimensional subspaces that can be optimized separately. \cite{mutny2018efficient} extend the approach by allowing overlapping subspaces. However, these methods rely on  hand-crafted assumptions that may be violated in most real-world applications.

The most recent approaches take advantage of deep generative models to both reduce dimensionality and take advantage of the structure of the latent space. These approaches assume access to a reservoir of unlabelled data that can be used in learning a nonlinear, low dimensional manifold from the data in an unsupervised way. \cite{griffiths2017constrained} use a two-step approach by first training an Auto Encoder (AE) with all unlabelled data to find a low dimensional representation of the problem and then performing BO in the low-dimensional latent space. \cite{kusner2017grammar} extend the approach to better handle the uncertainty by resorting to deep generative models. In particular, rather than using an AE in the first stage, they resort to a Variational Auto Encoder (VAE) to model the uncertainty to the latent space. In the second stage, they use a Gaussian Process Latent Variable Model (GPLVM) to propagate the uncertainty of the latent space improving the overall performance.

The biggest problem with these two-step approaches is that the latent space learned while only using the unlabelled data might not be optimal for the optimization task. \cite{eissman2018bayesian} address this issue by jointly learning the latent space with the labelled data and iteratively modifying the latent representation as new data is being collected. \cite{tripp2020sample} further exploit the retraining of the latent space by weighing the samples used in training the VAE based on their observed values so that good observations have more importance in training the latent space.

The approaches combining deep generative models with Gaussian Processes seem the most prominent approach to the problem at the moment. The strength of the approach is in its ability to use the unlabelled data together with the labelled data. The use of the unlabelled data allows the use of non-linear manifolds for optimization and is thus suitable for complex real world problems. This is the reason why in this paper we concentrate on these methods. Specifically, we use VAEs as deep generative models due to them having become the de facto method for the problem. The popularity of VAEs has led to them being applied on different types of structured data, including but not limited to images \citep{hou2017deep}, text \citep{pmlr-v70-yang17d}, sound \citep{46809} and structured physical objects like molecules \citep{gomez2018automatic}. This allows practitioners from many fields of work to benefit from our work. In the remaining of the paper, we analyze the effect of different design choices in the optimization approaches based on VAEs. There is a need for this kind of work as these approaches appear very promising to the practitioners, but the existing research does not yet study the effect of different design choices when applying these methods in practice.

\section{Theory and derivations}
\label{sec:theory}
In this section we introduce all necessary parts to define the general framework we will use for the analysis of Bayesian Optimization using deep generative models. The section first introduces  Variational Auto Encoders (VAEs), then Gaussian Latent Variable Models (GPLVMs) and finally how to combine VAEs and GPLVMs.

\subsection{Variational Auto Encoders}
In Variational Auto Encoders (VAEs) the aim is to find a latent representation $\z \in \mathbb{R}^d$ for data $\x\in \mathcal{X}$. Let $\theta$ parametrize a probabilistic decoder $p_\theta ( \x \given \z)$ with prior distribution $p(\z)$. The posterior distribution $p_\theta (\z \given \x) \propto p_\theta (\x \given \z ) p(\z)$ can be interpreted as a probabilistic encoder, which in most cases is intractable. In order to address this, the posterior needs to be approximated with a tractable distribution $q_\phi (\z \given \x)$, where $\phi$ parametrizes the encoder. The parameters $\theta$ and $\phi$ can jointly be learnt by maximizing the evidence lower bound (ELBO)
\begin{equation}
\begin{split}
\mathcal{L}(\phi, \theta ; \x) = & \E_{q_\phi (\z |\x)} \left[\log p_\theta(\x | \z) \right]  \\ &- KL \left(q_\phi (\z | \x) || p(z)\right), \label{eq:elbo}
\end{split}
\end{equation}
which can be done with gradient descent  as long as decoder and encoder approximation are differentiable with respect to $\theta$ and $\phi$ and are computable pointwise. A normal choice for the encoder distribution is multivariate normal, $q_\theta(\z \given \x) = \Normal{\z \given \boldsymbol{\mu}_\theta (\x), \boldsymbol{\Sigma}_\theta (\x)}$, where mean and (usually) diagonal covariance are outputs of a neural network. Decoder distributions vary more, depending on th type of the data. Bernoulli distributions can be used to decode binary data, continuous Bernoulli for bounded data (\cite{loaiza2019continuous}) and (log) normal distribution for (half-bounded) continuous data.

\subsection{Gaussian Process Latent Variable Models \label{sec:gvae1}}
Traditional BO approaches assume that the true latent black-box function $f$ is a realization of random variables sampled from a Gaussian process, $p(f) = \mathcal{GP}$ fully specified by a prior mean and some covariance function $K$ \citep{rasmussen2003gaussian}. The prior mean, which is often zero, defines the prior mean of the latent function and the covariance function specifies the covariance of the latent function between any two points. The problem with full GPs is their scalability. Assuming $N$ observations from the latent function, computing the posterior of a full GP requires inverting a $N\times N$ matrix.

GPs can be approximated, and made more scalable, by using inducing inputs. In this method we use inducing latent values $\iu$ (at locations $\z_u$) in the latent space instead of latent values $\f$ with observations $\y$ (at input locations $\z$). Using this notation, the posterior of the data becomes
\begin{equation}
p(\y \given \iu, \z_u, \z) = \E_{p(\f \given \iu)} \left [  p(\y \given \f) \right].
\end{equation}
The posterior of the inducing latent values, $p(\iu \given y, \z_u, \z) \propto \int p(\iu \given \f, \z_u, \z) p(\f \given \y, \z) \id \f \propto \int p(\iu \given \f, \z_u, \z) p(\y \given \f) p(\f \given \z) \id \f $, is intractable for general likelihood $p(\y \given \f)$ and for efficient prediction needs to be approximated. Approximating it with normal $q(\iu) = \Normal{ \iu \given \am, \aL^\top \aL} $, with general mean and general lower triangular matrix provides computationally tractable properties.

The GP log posterior likelihood can be approximated with a lower bound
\begin{equation}
\log p(\y ) \geq  \E _{q(\iu)} \left [ \log p(\y \given \iu) \right]   - \mathrm{KL} \left[ q(\iu) \; | | \; p(\iu) \right],
 \label{eq:elbogplvm}
\end{equation} 
where $p(\iu) = \Normal{\iu \given \bs{\mu}_{prior}, \bs{\Sigma}_{prior}}$, with $\bs{\mu}_{prior}$ and $\bs{\Sigma}_{prior}$, which are the \gp prior mean and prior covariance. Furthermore, with $p(y \given f) = \Normal{y \given f, \sigma^2}$, the likelihood inside the expectation in Equation \eqref{eq:elbogplvm} reduces as
\begin{align}
&p(\y \given \iu ) & = &\int p(\y \given \f) p(\f \given \iu) \id \f \nonumber\\
&&= &\Normal{\f \given \mathbf{A} \mathbf{m}, \mathbf{K}_{ff} + \mathbf{A} ( \aL^\top \aL - \mathbf{K}_{uu}) \mathbf{A} ^ \top}, \label{eq:uposterior}
\end{align}
where $\mathbf{A} = \mathbf{K}_{fu} \mathbf{K}_{uu}^{-1}$. This model is referred to as Stochastic Variational Gaussian process (\svgp, \cite{snelson2005sparse,pmlr-v5-titsias09a}).

Since the latent space of a VAEs is uncertain, we can also add uncertainty to the locations $\z$ by assuming that the projections in the latent space follow an unknown distribution $q_\phi(\z \given \x) = \Normal{\z \given \mu_\phi(\x), \Sigma_\phi(\x)}$, parametrized by the encoder. This uncertainty can be added to the lower bound as 
\begin{eqnarray}
\log p(\y \given \z)  \geq \E _{q(\f, \z)} \left [ \log p(\y \given \f) \right] - \mathrm{KL} \left[ q(\f) \; | | \; p(\iu) \right] \nonumber \\   = \E _{q(\z)} \left [ \E _{q(\f \given \z)} \left [ \log p(\y \given \f) \right] - \mathrm{KL} \left[ q(\f) \; | | \; p(\iu) \right] \right] \label{eq:gp}.
\end{eqnarray}
This model, GP with uncertainty in the inputs, is called Gaussian Process Latent Variable Model (GPLVM) (\cite{lawrence2004gaussian}).

\subsection{Gaussian Processes in the latent space of Variational Auto Encoder \label{sec:GPs}}
To the best of our knowledge, there are two widely used alternatives for combining GPLVM with VAE. For general notation, let $\{\X_o, \Y_o\}$ be the observed data and observations and $\X_u$ be the unobserved data.
\subsubsection{Training VAE and GPLVM disjointly} \label{sec:dissjointly}
The original and simplest way of combining GPLVM with VAE was introduced by \cite{kusner2017grammar}. Their approach is to train the VAE with $\X_u$ prior to any observations using equation \eqref{eq:elbo}. After training the VAE, its parameters are fixed and the GPLVM is trained by maximizing equation \eqref{eq:gp} with $p_\phi(\Z_o \given \X_o)$, where $\phi$ is fixed.

\subsubsection{Training VAE and GPLVM jointly}
An alternative approach is to train the parameters of the VAE and GPLVM jointly like \cite{eissman2018bayesian}. Their approach uses separate cost functions for the labelled and unlabelled data. For labelled data costs of equations \eqref{eq:elbo} and \eqref{eq:gp} are combined as,
\begin{equation}
\begin{split}
\mathcal{L}(\phi, \theta; \x_o, \y_o) = \mathcal{L}(\phi, \theta; \x_o) \\
+ \E_{q_\phi (\z_o \given \x_o)} \left [ \log p(\y_o \given \z_o) \right].
\end{split}
\end{equation}
For unlabelled data, the cost defined in equation \eqref{eq:elbo} is used.

\section{Bayesian optimization with variational auto encoders}\label{sec:bovae}
Bayesian Optimization (BO) is a gradient-free black-box optimization method. The iterative steps of any BO algorithm are: 1) Train the probabilistic surrogate model, usually a Gaussian process or in our case GPLVM, using the available data; 2) Evaluate the black-box function at the maximum of the acquisition function; 3) Update the existing data with new evaluation; 4) Repeat steps 1, 2 and 3 until a certain stopping criterion is met. When dealing with high dimensional structured spaces, the first step also includes (re)-training the latent space either jointly or disjointly with the surrogate model. Also, the optimization bounds of the acquisition function need to be re-learned on every iteration as the mapping to the latent space constantly changes (see algorithm \ref{alg:sbo}). In this section we discuss different design choices for applying BO to tasks in high dimensional structured spaces: 1) the dimensionality of the latent space, 2) the choice of the acquisition function and 3) the choice of the optimization bounds in the latent space.

\begin{algorithm}[tb]
   \caption{Bayesian Optimization with variational auto encoders.}
   \label{alg:sbo}
\begin{algorithmic}[1]
   \STATE {\bfseries Input:} Unlabelled data $\X_{u}$, labelled data $\left\{ \X_o, \Y_o\right\}$, acquisition function $A(\cdot)$, black-box function $f(\cdot)$
   \REPEAT
   \STATE (Re-)Learn the encoder, decoder and GP parameters $\theta, \phi$ using $\X_{u}, \X_o$ and $\Y_o$
   \STATE (Re-)Learn the space $\mathcal{Z} \in \mathbb{R}^d$ where the acquisition function is optimized using $\X_u$, $\theta$ and $\phi$
   \STATE Find the next location $\z_*$ in the latent space by maximizing the acquisition function $\z_* = \argmax _{\z \in \mathcal{Z}} A(\z)$
   \STATE Project $\z_*$ to the original data space as $\x_*$ using the learned decoder parameters $\theta$
   \STATE Find label $y_*$ using $f(\x_*)$
   \STATE Append $\left\{ \X_o, \Y_o\right\}$ with $\left\{ \x_*, y_*\right\}$
   \UNTIL{Evaluation budget is over or acquisition is lower than threshold}
\end{algorithmic}
\end{algorithm}
\subsection{The dimensionality of the latent space}\label{sec:resultsdim}
Choosing the right dimensionality of the latent space is complex and task dependent. A too low dimensional latent space affects the quality of the samples in $\mathcal{X}$ produced by the decoder. On the other hand, a too high dimensional latent space makes fitting the GPLVM in the latent space harder and not as sample efficient. In addition, a too high dimensional space leads to overfitting and poor generalization. The aim of iteratively learning the latent space with the collected observations is to make the optimization task easier in the latent space, but it is yet an open research question how the methods perform when the latent space dimensionality is varied.

\subsection{The choice of acquisition function}
It has been deeply studied how different acquisition functions perform in low dimensional Euclidean spaces that are prevalent in traditional BO applications. All acquisition functions balance between exploration and exploitation; the tendency of sampling from regions with lots of uncertainty versus tendency of sampling from regions with known good values. Here we explore the role of the acquisition in structured high-dimensional spaces. In particular, the acquisition functions used in this paper include Thompson Sampling (TS) (\cite{thompson1933likelihood, NIPS2011_4321}), Expected Improvement (EI), Probability of Improvement (PI) and Lower Confidence Bound (LCB). (\cite{snoek2012practical}) 

\subsection{Optimization bounds of the acquisition function}
Unlike in the regular low dimensional BO, the selection of the optimization bounds of the acquisition function is a difficult design choice. Normally optimization bounds for the acquisition function are selected based on expert knowledge or physical constraints. As the latent space formed by the VAE is an abstraction and as such is not tied to the business domain of the problem at hand, selecting the bounds for it is much harder. So far, the de facto method has been to bound the optimization of the acquisition function by a hypercube containing the projection of the training data in the latent space. To the best of our knowledge, no alternatives to this method have been explored.

We demonstrate three methods for restricting the optimization space. An easy and scalable approximation is to find a minimum volume n-ellipsoid, $(\x-\x_0)\tr \mathbf{A} (\x-\x_0)=1$, that contains (the means of) all the training data. An easy and scalable way of doing this is via the Khachiyan algorithm \cite{moshtagh2005minimum}. Another easy, but not as scalable way is to find a set of hyperplanes restricting the data and form a set of linear inequalities of the form $\mathbf{A} \x \preceq \mathbf{b}$. It is easy to find this set of inequalities by first finding a convex hull for the existing samples and then perform Delaunay triangulation for the edge points and use the edge points of each simplex to find the hyperplanes. The combined time complexity of these operations is $\mathcal{O}(n^2)$, where $n$ is the total number of unlabelled points. The benefit of both these methods is that the optimization of the acquisition function can be performed in a convex set.

Third approach of limiting the optimization space is setting an upper limit to the allowed distance between a point in the latent space $\z'$ and the expected value of the encoder of the expected value of the decoded $\z'$
\begin{equation}
\left|\left|\z' - \int \z p_\theta \left(\z \middle | \int \x q_\phi\left(\x \middle | \z \right) \id \x \right) \id \z \right|\right|.
\end{equation} 

\begin{figure}[tb!]
    \centering
    \includegraphics{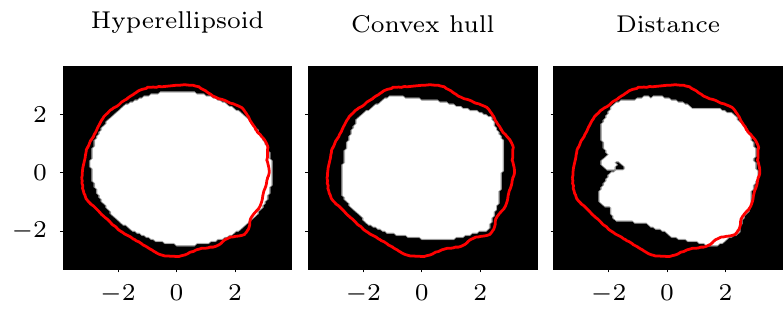}
    \vspace{-3mm}
\caption{Different space restriction strategies for acquisition function optimization visualized from left to right ellipse, convex hull and distance method. In the distance method the maximum distance the point can move is computed as a 90\% of the maximum of the move distances of latent space means of the train data. The white regions indicate the area where the acquisition function is optimized. The red continuous line visualizes the region which contains 99\% of the data used to train the VAE.  }
\label{fig:restrictions}
\end{figure}

 This approach guarantees that the uncertainty is reduced in the regions where the acquisition function is originally meant to be evaluated. A good strategy for selecting the upper bound of the Euclidean distance is to see how much all the points in the training data move and select e.g. the 90\% percentile of these distances. The drawback of the approach is that the allowed region is not necessarily convex, making the optimization harder. All these strategies are visualized in figure \ref{fig:restrictions} for a VAE trained with the Shape dataset (see details of the Shape data set in section \ref{sec:results}).

\section{Experimental results}\label{sec:results}

\subsection{Experimental Setup}

All case studies are performed with the two variations of the high dimensional structured BO as described in \ref{sec:GPs}. We used MXNet (\cite{chen2015mxnet}) for modeling and Emukit (\cite{paleyes2019emulation}) to run the customized Bayesian Optimization routine. All the code written to run the experiments is fully available at \url{https://github.com/esiivola/hdssbo}.

The VAE used in the experiments has 3 layers in both encoder and decoder with [\{num inputs\}, \{num inputs\}, \{dim latent space\}$\times 2$] units in the encoder and [\{dim latent space\}, \{num inputs\}, \{num inputs\}] units in the decoder. The parameters are trained with a learning rate $10^{-3}$. The GPLVM model has 150 inducing points and a squared exponential kernel with each latent dimension having its own length scale parameter. Learning rate $10^{-1}$ is used for the GPLVM parameters. All parameters are trained using 'Adam'-algorithm \cite{kingma2014adam}. Prior to starting the optimizations routine, the GPLVM model is initialized with 10 observations sampled uniformly at random from the training data. Following \cite{tripp2020sample}, the VAE parameters are allowed to change only every $10$ iterations. The purpose of this is to save computation time and reduce overfitting.

\subsection{Data sets}
We design three structured high-dimensional optimization task based on the following datasets:

\textbf{Airline passenger dataset} is a time-series data consisting of the number of monthly airline passengers from January 1949 to December 1960 \cite{box2011time}. The black-box function is the mean square error of a model fitted on 66\% of the data on a test set consisting of 33\% of the data. Following the experimental setting in \cite{lu2018structured}, the fitted model is a Gaussian Process whose kernel is generated by a grammar with four basis kernels (periodic, squared exponential, linear and rational quadratic) and two operators (+ and *) so that there are at maximum 4 kernel combined in the produced kernel.

\textbf{2d shape area maximization dataset} is an image dataset consisting of rotated black rectangles of a varying area on a background of $10$ by $10$ pixels. The black-box function is the area of the rotated shape. The dataset mimics the dSprites dataset (\cite{dsprites17}) and is used for optimization in \cite{tripp2020sample}. The dataset simulates a situation, where the true optimum is not inside the training data and finding the optimum requires modifying the latent space.

\textbf{Molecule dataset} consists of valid molecules of carbon, oxygen, nitrogen and iron each containing up to 7 atoms in total and was first introduced by \cite{fink2007virtual}. The molecules in the dataset are described by the SMILES grammar (\cite{weininger1988smiles}). SMILES molecules can be embedded in a low dimensional space using a VAEs as described by \cite{kusner2017grammar}. The black-box function to be optimized is the penalized water-octanol partition coefficient that mimics the drug-likeliness of a molecule \cite{ertl2009estimation} and is also used in \cite{kusner2017grammar}. 

\subsection{Effect of the dimensionality of the latent space}
\begin{figure}[tb!]    
    \centering
    \includegraphics{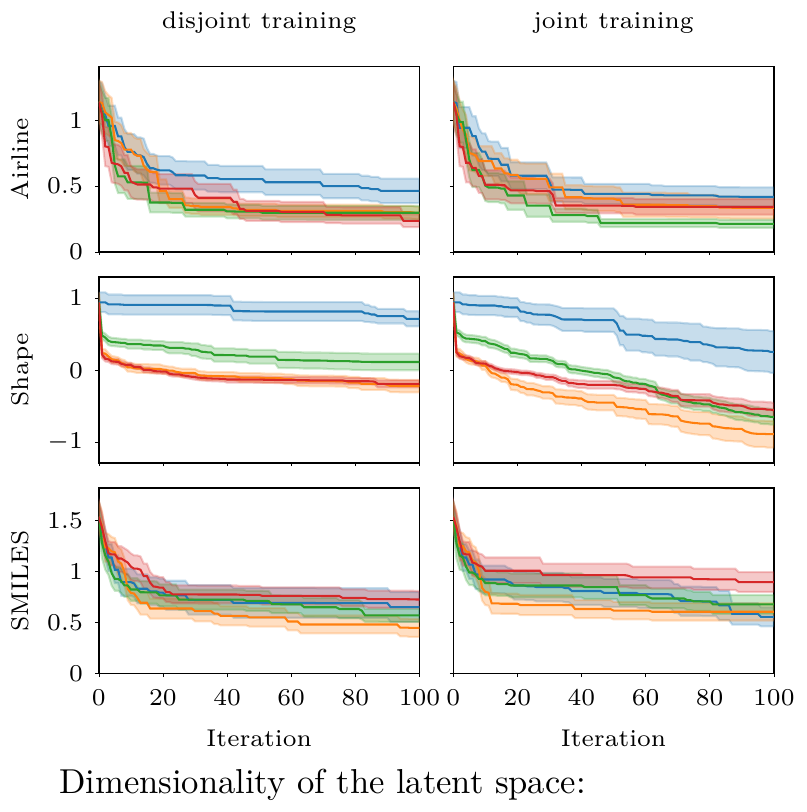}\\
    \noindent\\
    \vspace{-1.7mm}
    \hbox{\hspace{10mm}
	\includegraphics{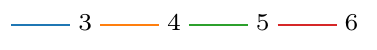}}
	\vspace{-3mm}
\caption{The effect of the latent space dimensionality on the performance of the BO algorithm. Each subplot visualizes the best observed value so far as a function of optimization iterations for different latent space dimensionalities. Each line visualizes the mean of ten separate runs and the colored band around the solid line visualizes the standard deviation of the mean. Different rows show results for different datasets and columns show results for different methods. The black-box function values are normalized so that the best value in the training data is 0 and the standard deviation of the values of the training data is 1. } \label{fig:dims}
\end{figure}

To study the robustness of changing the dimensionality of the latent space $d$, we compare the optimization performance on the three datasets and two models when trying different dimensionalities $d \in \{3, 4, 5, 6\}$. The acquisition space is restricted by a hypercube and the acquisition function is set to the Lower Confidence Bound. The results are visualized in figure \ref{fig:dims}.
The results show a clear trend for all datasets and both methods. For Airline, Shape and SMILES the best performance is obtained with 4 dimensional latent space. This is true for both tested method, but for the Shape dataset, at later iterations 5 dimensional latent space performs better than 4 dimensional latent space. The results also show that if the dimensionality is increased, the performance increases until it plateaus or starts to decrease again. This is caused by one one hand the optimization problem becoming harder in high dimensions but on the other hand the latent space becoming more nuanced with more dimensions. The first property makes BO harder and the second property reduces how noisy the black-box function appears in the latent space.

The only dataset for which the joint training method outperforms the disjoint method is the Shape-dataset. The reason for this is that the joint training often leads to overfitting, which reduces its performance. However, the Shape dataset is optimal for joint training as the minimum of a black-box function is outside the data used to train the VAE. In other words, excellent performance requires big changes in VAE

\subsection{Effect of different optimization space selection strategies}
\begin{figure}[tb!]
    \centering
    \includegraphics{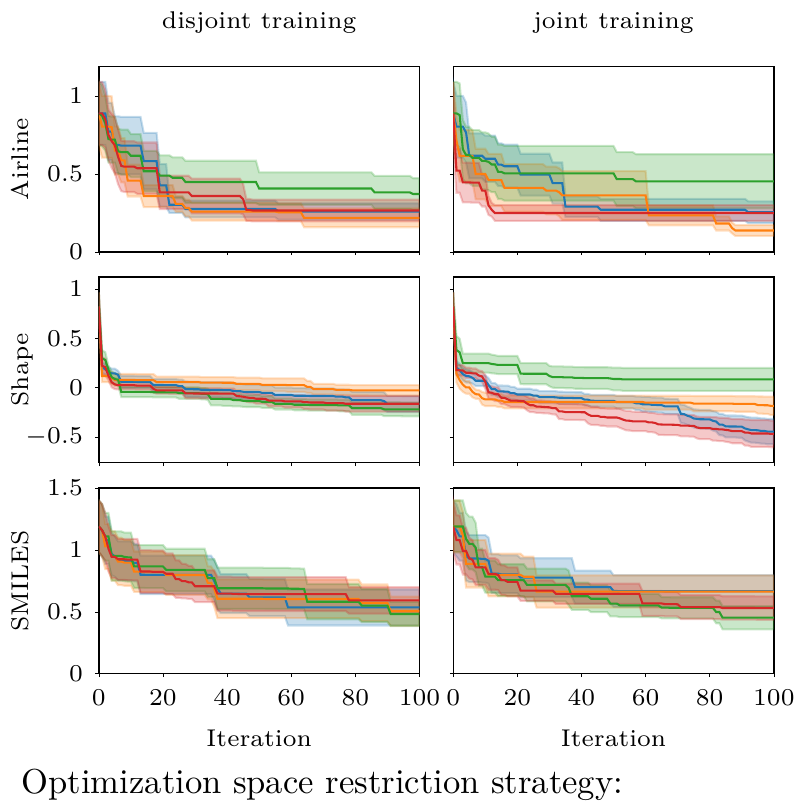}
	\vspace{-1mm}	
	\includegraphics{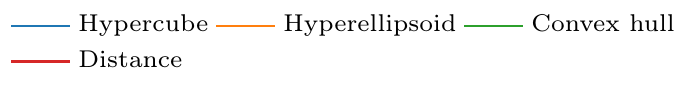}
	\vspace{-3mm}
\caption{The effect of using different strategies of defining the optimization space on the performance of the BO algorithm. Representation of the lines, colored bands and normalized black-box values is the same as in figure \ref{fig:dims}.} \label{fig:space}
\end{figure}

To study the effect of different acquisition space optimization strategies, we compare the optimization performance of the three datasets and two models using the three acquisition space restriction strategies described in Section \ref{fig:restrictions}, i.e. the hyperellipsoid, minimum convex hull and metric based on minimizing the extrapolation error. In addition to these, the simple hyper rectangle approach traditionally used in the BO literature is used as a baseline. The results are visualized in Figure \ref{fig:space}.

\begin{figure}[tb!]
    \centering
    \begin{tabular}{p{0.31\textwidth}  p{0.1\textwidth}}
    \vspace{0pt} \includegraphics{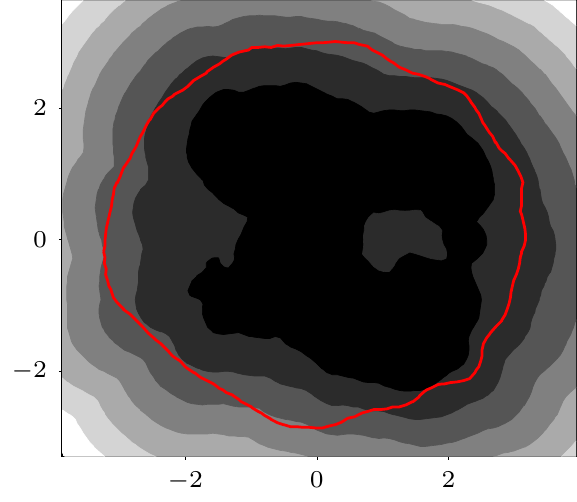} &  \vspace{0pt} \begin{tabular}{l} \includegraphics{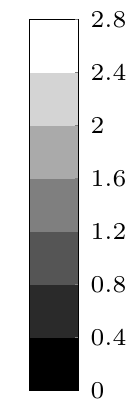} \\ \includegraphics{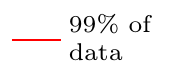} \end{tabular} \\
    \end{tabular}
\vspace{-4mm}
\caption{The Euclidean distance between a point $\z'$ in a latent space and the location of $\mu_\theta(\x')$ (where $\x'$ is the decoded $\z'$) as a function of $\z'$ in two dimensional latent space. The red continuous line visualizes the region which contains 99\% of the data used to train the VAE projected in the latent space of the encoder. Darker colors mean shorter distance and brighter colors mean larger distance. The VAE which latent space is visualized is trained using the Shape-data.}
\label{fig:extrapolation}
\end{figure}

The results are surprising as different optimization space  restriction strategies seem to only have a very minimal effect on the results. To understand why we need to first understand how the VAE works on the edges of the search space. If a point $\z' \in \mathbb{R}^d$ is selected from the corner of the hypercube in the latent space, it needs to be projected to $\int \x q_\phi(\x \given \z') \id \x =  \x' \in \mathcal{X}$ to be evaluated. As the corner of the hypercube is outside the data (or on the edge if we are lucky), the decoder needs to extrapolate as it has not been trained using this kind of data. After evaluation, for $\x'$ to be usable for the GPLVM, it needs to be projected back to the latent space $\mathbb{R}^d$ (as a distribution $p_\theta(\z^*\given \x')$). Since the decoder extrapolates, the point projected back to the latent space is not the same as $\z'$. figure \ref{fig:extrapolation} visualizes the Euclidean distance between the original point in the latent space $\mathcal{X}$ and the point that has first been projected to the original space $\x' = \int \x p(\x \given \z') \id \x$ and then back to the latent space $\z^* = \int \z^* q_\phi(\z \given \x') \id \z$. The figure shows that the further away the point is from the data used to train the VAE, the bigger the distance is. The figure does not show it, but the points sampled from outside the training data travel closer to the training data.

\begin{figure}[tb!]
    \centering
    \includegraphics{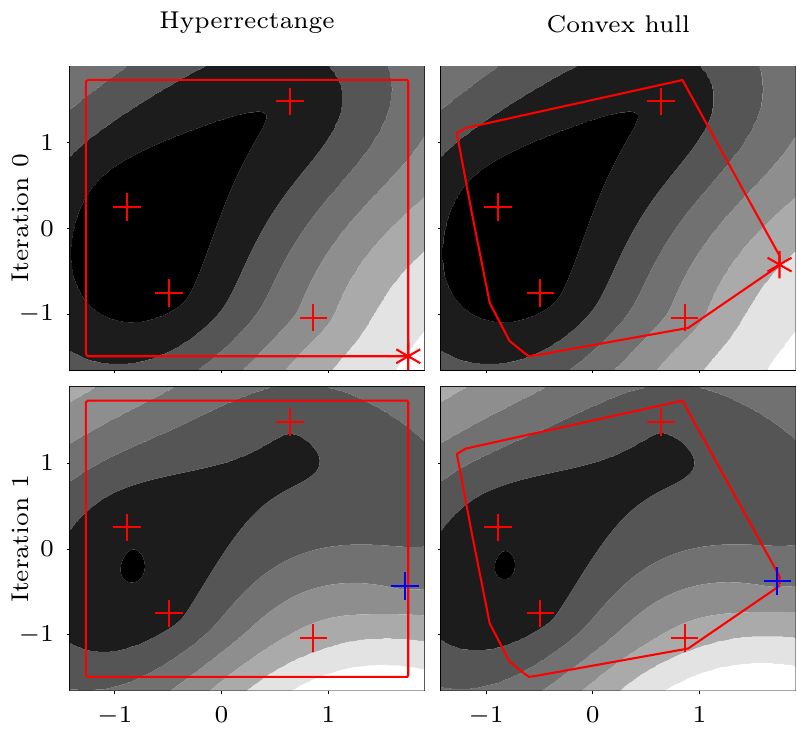}
    \includegraphics{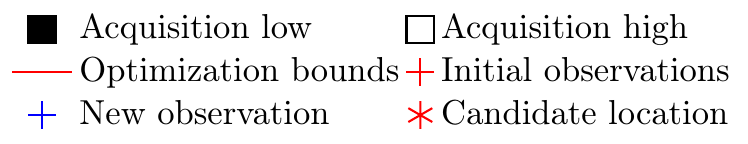}
    \vspace{-3mm}
\caption{The figure shows, on the example of the Airline dataset, how the acquisition function changes during one iteration in a BO for two separate optimization techniques (columns) a) space bounded with a hyper rectangle bounding the training data b) space bounded with a smallest convex hull bounding the training data. On iteration 0, the only difference between the two techniques is the location of the maximum of the acquisition function (marked as {\color{red}*}). Even though the acquisition function surfaces are identical the maximums within the optimization area are different. On iteration 1, the maximum of the previous iteration has been projected to the original space to be evaluated and the evaluated point has been projected back to the latent space (marked as {\color{blue}+}). Since for the hyperrectangle approach, the maximum of the acquisition function is outside the data used to train the VAE, the distance between the maximum of the acquisition function ({\color{red}*}) at iteration 0 and the evaluated point ({\color{blue}+}) at iteration 1 is large. As the maximum of the acquisition function is within the training data for the convex hull approach, the distance is much smaller. Since the maximums of the acquisition function are close to each other at iteration 0 for both restriction strategies, the points are evaluated at the same location and both methods have identical outcome after the first iteration.} \label{fig:poor_extrapolation}
\end{figure}

The poor extrapolation of VAE also changes how the BO works. Figure \ref{fig:poor_extrapolation} visualizes this on the airline dataset with the acquisition function LCB. The figure visualizes how the acquisition function looks in the latent space of the VAE using two different optimization space restriction strategies in different phases of the BO loop. As the optimization space restriction methods are different,  the maximums of the acquisition functions are in different locations. However, for the hyperrectangle approach, as the point is outside the training data, the decoder has to extrapolate and when the extrapolated and evaluated point is projected back to the latent space, it has traveled towards the data set (the distance between the red '*' on the first row and blue '+' on the second row). When comparing to the convex hull approach, the evaluated point has not traveled that much. After all, the effect of both methods is very similar. Counter intuitively, the poor extrapolation of the VAE causes similar behaviour as more sophisticated optimization space restriction strategies.

\subsection{Effect of different acquisition strategies}
\begin{figure}[tb!]
    \centering
    \includegraphics{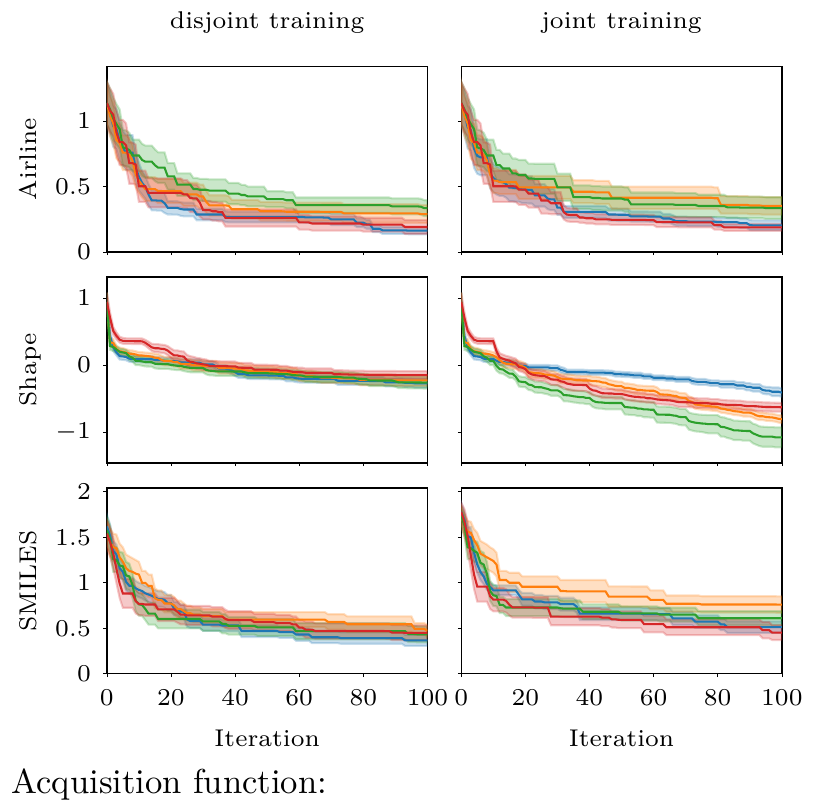}
	\includegraphics{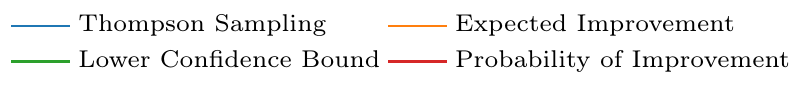}
	\vspace{-7mm}
\caption{The effect of using different acquisition functions on the performance of the BO algorithm. Representation of the lines, colored bands and normalized black-box values is the same as in figure \ref{fig:dims}.} \label{fig:acqs}
\end{figure}

To study the effect of different acquisition functions, we compare the optimization performance of the three datasets and two models using four different acquisition functions using four dimensional latent space. The results are visualized in figure \ref{fig:acqs}.

The different acquisition strategies show a behaviour that is comparable to their expected behaviour in regular, low dimensional, BO settings. This means that there is no single best acquisition strategy that would rule them all. For Airline data, exploitative strategies EI and PI perform better than the explorative strategies. For Shape data, explorative strategies perform better than the exploitative strategies, which is natural as the minimum is outside the training data. For SMILES data, explorative strategies perform slightly better. There is no significant difference between different methods for the same acquisition function. The order of the performance is similar for both the tested methods. Also, as demonstrated in the previous experiments, the performance of the joint training for Airline and SMILES datasets is affected by overfitting.

\section{Conclusion}\label{sec:conclusion}

In this paper, we demonstrated how to design an optimization task in high dimensional structured spaces. We concentrated on methods combining deep generative models, such as Variational Auto Encoders, and Gaussian processes so that the regression model and optimization are performed in the smooth Euclidean low dimensional space found by the generative model.

We demonstrated the effect of dimensionality, optimization space restriction strategy and acquisition functions on the optimization performance. The results show that there is an optimum dimension for the deep generative model in which the optimization yields the best performance. We also demonstrated that there is no need to restrict the optimization space of the acquisition function. In addition to this, we demonstrated that acquisition functions have similar performance in structured problems as they do in the case of regular, low dimensional BO.

We also demonstrated that tuning the latent space by jointly learning the latent space and the GP model might lead to overfitting and an overall drop in performance. This can be seen in the case of Airline and Shape datasets, where the disjoint training approach performed better.

The remaining open question and potential topic for further research is how to avoid the apparent overfitting of the VAE parameters when training the VAE and GPLVM models jointly. Another topic for further research is to use sparse Variational Auto Encoders (\cite{tonolini2020variational}) in BO to avoid the problem of having to decide the optimal latent space dimensionality (as demonstrated in Section \ref{sec:resultsdim}).

\subsection*{Acknowledgements}
The authors would like to thank Pablo Garc{\'i}a Moreno for his valuable feedback and discussions that helped us in improving the manuscript. The authors thank Austin Tripp for improving the clarity of the manuscript.

\bibliography{vpbbo}

\begin{thebibliography}{33}
\providecommand{\natexlab}[1]{#1}
\providecommand{\url}[1]{\texttt{#1}}
\expandafter\ifx\csname urlstyle\endcsname\relax
  \providecommand{\doi}[1]{doi: #1}\else
  \providecommand{\doi}{doi: \begingroup \urlstyle{rm}\Url}\fi

\bibitem[Box et~al.(2011)Box, Jenkins, and Reinsel]{box2011time}
George~EP Box, Gwilym~M Jenkins, and Gregory~C Reinsel.
\newblock \emph{Time series analysis: forecasting and control}, volume 734.
\newblock John Wiley \& Sons, 2011.

\bibitem[Brochu et~al.(2010)Brochu, Cora, and De~Freitas]{brochu2010tutorial}
Eric Brochu, Vlad~M Cora, and Nando De~Freitas.
\newblock A tutorial on {B}ayesian optimization of expensive cost functions,
  with application to active user modeling and hierarchical reinforcement
  learning.
\newblock \emph{arXiv preprint arXiv:1012.2599}, 2010.

\bibitem[Chapelle and Li(2011)]{NIPS2011_4321}
Olivier Chapelle and Lihong Li.
\newblock An empirical evaluation of {T}hompson sampling.
\newblock In J.~Shawe-Taylor, R.~S. Zemel, P.~L. Bartlett, F.~Pereira, and
  K.~Q. Weinberger, editors, \emph{Advances in Neural Information Processing
  Systems 24}, pages 2249--2257. 2011.

\bibitem[Chen et~al.(2015)Chen, Li, Li, Lin, Wang, Wang, Xiao, Xu, Zhang, and
  Zhang]{chen2015mxnet}
Tianqi Chen, Mu~Li, Yutian Li, Min Lin, Naiyan Wang, Minjie Wang, Tianjun Xiao,
  Bing Xu, Chiyuan Zhang, and Zheng Zhang.
\newblock {MXN}et: A flexible and efficient machine learning library for
  heterogeneous distributed systems.
\newblock \emph{arXiv preprint arXiv:1512.01274}, 2015.

\bibitem[Eissman et~al.(2018)Eissman, Levy, Shu, Bartzsch, and
  Ermon]{eissman2018bayesian}
Stephan Eissman, Daniel Levy, Rui Shu, Stefan Bartzsch, and Stefano Ermon.
\newblock Bayesian optimization and attribute adjustment.
\newblock In \emph{Proc. 34th Conference on Uncertainty in Artificial
  Intelligence}, 2018.

\bibitem[Ertl and Schuffenhauer(2009)]{ertl2009estimation}
Peter Ertl and Ansgar Schuffenhauer.
\newblock Estimation of synthetic accessibility score of drug-like molecules
  based on molecular complexity and fragment contributions.
\newblock \emph{Journal of cheminformatics}, 1\penalty0 (1):\penalty0 8, 2009.

\bibitem[Fink and Reymond(2007)]{fink2007virtual}
Tobias Fink and Jean-Louis Reymond.
\newblock Virtual exploration of the chemical universe up to 11 atoms of {C},
  {N}, {O}, {F}: assembly of 26.4 million structures (110.9 million
  stereoisomers) and analysis for new ring systems, stereochemistry,
  physicochemical properties, compound classes, and drug discovery.
\newblock \emph{Journal of chemical information and modeling}, 47\penalty0
  (2):\penalty0 342--353, 2007.

\bibitem[Garnett et~al.(2013)Garnett, Osborne, and Hennig]{garnett2013active}
Roman Garnett, Michael~A Osborne, and Philipp Hennig.
\newblock Active learning of linear embeddings for gaussian processes.
\newblock \emph{arXiv preprint arXiv:1310.6740}, 2013.

\bibitem[G{\'o}mez-Bombarelli et~al.(2018)G{\'o}mez-Bombarelli, Wei, Duvenaud,
  Hern{\'a}ndez-Lobato, S{\'a}nchez-Lengeling, Sheberla, Aguilera-Iparraguirre,
  Hirzel, Adams, and Aspuru-Guzik]{gomez2018automatic}
Rafael G{\'o}mez-Bombarelli, Jennifer~N Wei, David Duvenaud, Jos{\'e}~Miguel
  Hern{\'a}ndez-Lobato, Benjam{\'\i}n S{\'a}nchez-Lengeling, Dennis Sheberla,
  Jorge Aguilera-Iparraguirre, Timothy~D Hirzel, Ryan~P Adams, and Al{\'a}n
  Aspuru-Guzik.
\newblock Automatic chemical design using a data-driven continuous
  representation of molecules.
\newblock \emph{ACS central science}, 4\penalty0 (2):\penalty0 268--276, 2018.

\bibitem[Griffiths and Hern{\'a}ndez-Lobato(2017)]{griffiths2017constrained}
Ryan-Rhys Griffiths and Jos{\'e}~Miguel Hern{\'a}ndez-Lobato.
\newblock Constrained {B}ayesian optimization for automatic chemical design.
\newblock \emph{arXiv preprint arXiv:1709.05501}, 2017.

\bibitem[Hou et~al.(2017)Hou, Shen, Sun, and Qiu]{hou2017deep}
Xianxu Hou, Linlin Shen, Ke~Sun, and Guoping Qiu.
\newblock Deep feature consistent variational autoencoder.
\newblock In \emph{2017 IEEE Winter Conference on Applications of Computer
  Vision (WACV)}, pages 1133--1141, 2017.

\bibitem[Kandasamy et~al.(2015)Kandasamy, Schneider, and
  P{\'o}czos]{kandasamy2015high}
Kirthevasan Kandasamy, Jeff Schneider, and Barnab{\'a}s P{\'o}czos.
\newblock High dimensional {B}ayesian optimisation and bandits via additive
  models.
\newblock In \emph{International conference on machine learning}, pages
  295--304, 2015.

\bibitem[Kingma and Ba(2014)]{kingma2014adam}
Diederik~P Kingma and Jimmy Ba.
\newblock Adam: A method for stochastic optimization.
\newblock \emph{arXiv preprint arXiv:1412.6980}, 2014.

\bibitem[Kusner et~al.(2017)Kusner, Paige, and
  Hern{\'a}ndez-Lobato]{kusner2017grammar}
Matt~J Kusner, Brooks Paige, and Jos{\'e}~Miguel Hern{\'a}ndez-Lobato.
\newblock Grammar variational autoencoder.
\newblock \emph{arXiv preprint arXiv:1703.01925}, 2017.

\bibitem[Lawrence(2004)]{lawrence2004gaussian}
Neil~D Lawrence.
\newblock Gaussian process latent variable models for visualisation of high
  dimensional data.
\newblock In \emph{Advances in neural information processing systems}, pages
  329--336, 2004.

\bibitem[Loaiza-Ganem and Cunningham(2019)]{loaiza2019continuous}
Gabriel Loaiza-Ganem and John~P Cunningham.
\newblock The continuous {B}ernoulli: fixing a pervasive error in variational
  autoencoders.
\newblock In \emph{Advances in Neural Information Processing Systems}, pages
  13287--13297, 2019.

\bibitem[Lu et~al.(2018)Lu, Gonzalez, Dai, and Lawrence]{lu2018structured}
Xiaoyu Lu, Javier Gonzalez, Zhenwen Dai, and Neil Lawrence.
\newblock Structured variationally auto-encoded optimization.
\newblock In \emph{International Conference on Machine Learning}, pages
  3267--3275, 2018.

\bibitem[Matthey et~al.(2017)Matthey, Higgins, Hassabis, and
  Lerchner]{dsprites17}
Loic Matthey, Irina Higgins, Demis Hassabis, and Alexander Lerchner.
\newblock d{S}prites: Disentanglement testing {S}prites dataset.
\newblock https://github.com/deepmind/dsprites-dataset/, 2017.

\bibitem[Moshtagh et~al.(2005)]{moshtagh2005minimum}
Nima Moshtagh et~al.
\newblock Minimum volume enclosing ellipsoid.
\newblock \emph{Convex optimization}, 111\penalty0 (January):\penalty0 1--9,
  2005.

\bibitem[Mutny and Krause(2018)]{mutny2018efficient}
Mojmir Mutny and Andreas Krause.
\newblock Efficient high dimensional {B}ayesian optimization with additivity
  and quadrature {F}ourier features.
\newblock In \emph{Advances in Neural Information Processing Systems}, pages
  9005--9016, 2018.

\bibitem[Paleyes et~al.(2019)Paleyes, Pullin, Mahsereci, Lawrence, and
  Gonzalez]{paleyes2019emulation}
Andrei Paleyes, Mark Pullin, Maren Mahsereci, Neil Lawrence, and Javier
  Gonzalez.
\newblock Emulation of physical processes with {E}mukit.
\newblock In \emph{Second Workshop on Machine Learning and the Physical
  Sciences, NeurIPS}, 2019.

\bibitem[Rasmussen(2003)]{rasmussen2003gaussian}
Carl~Edward Rasmussen.
\newblock Gaussian processes in machine learning.
\newblock In \emph{Summer School on Machine Learning}, pages 63--71. Springer,
  2003.

\bibitem[Roberts et~al.(2017)Roberts, Engel, and Eck]{46809}
Adam Roberts, Jesse Engel, and Douglas Eck, editors.
\newblock \emph{Hierarchical Variational Autoencoders for Music}, 2017.
\newblock URL
  \url{https://nips2017creativity.github.io/doc/Hierarchical_Variational_Autoencoders_for_Music.pdf}.

\bibitem[Shahriari et~al.(2015)Shahriari, Swersky, Wang, Adams, and
  De~Freitas]{shahriari2015taking}
Bobak Shahriari, Kevin Swersky, Ziyu Wang, Ryan~P Adams, and Nando De~Freitas.
\newblock Taking the human out of the loop: A review of {B}ayesian
  optimization.
\newblock \emph{Proceedings of the IEEE}, 104\penalty0 (1):\penalty0 148--175,
  2015.

\bibitem[Snelson and Ghahramani(2005)]{snelson2005sparse}
Edward Snelson and Zoubin Ghahramani.
\newblock Sparse gaussian processes using pseudo-inputs.
\newblock \emph{Advances in Neural Information Processing Systems 18},
  18:\penalty0 1257--1264, 2005.

\bibitem[Snoek et~al.(2012)Snoek, Larochelle, and Adams]{snoek2012practical}
Jasper Snoek, Hugo Larochelle, and Ryan~P Adams.
\newblock Practical {B}ayesian optimization of machine learning algorithms.
\newblock In \emph{Advances in neural information processing systems}, pages
  2951--2959, 2012.

\bibitem[Thompson(1933)]{thompson1933likelihood}
William~R Thompson.
\newblock On the likelihood that one unknown probability exceeds another in
  view of the evidence of two samples.
\newblock \emph{Biometrika}, 25\penalty0 (3/4):\penalty0 285--294, 1933.

\bibitem[Titsias(2009)]{pmlr-v5-titsias09a}
Michalis Titsias.
\newblock Variational learning of inducing variables in sparse gaussian
  processes.
\newblock In \emph{Proceedings of the 12th International Conference on
  Artificial Intelligence and Statistics (AISTATS)}, volume~5, pages 567--574.
  JMLR Workshop and Conference Proceedings, 2009.

\bibitem[Tonolini et~al.(2020)Tonolini, Jensen, and
  Murray-Smith]{tonolini2020variational}
Francesco Tonolini, Bj{\o}rn~Sand Jensen, and Roderick Murray-Smith.
\newblock Variational sparse coding.
\newblock In \emph{Uncertainty in Artificial Intelligence}, pages 690--700.
  PMLR, 2020.

\bibitem[Tripp et~al.(2020)Tripp, Daxberger, and
  Hern{\'a}ndez-Lobato]{tripp2020sample}
Austin Tripp, Erik Daxberger, and Jos{\'e}~Miguel Hern{\'a}ndez-Lobato.
\newblock Sample-efficient optimization in the latent space of deep generative
  models via weighted retraining.
\newblock \emph{Advances in Neural Information Processing Systems}, 33, 2020.

\bibitem[Wang et~al.(2013)Wang, Zoghi, Hutter, Matheson, De~Freitas,
  et~al.]{wang2013bayesian}
Ziyu Wang, Masrour Zoghi, Frank Hutter, David Matheson, Nando De~Freitas,
  et~al.
\newblock Bayesian {O}ptimization in high dimensions via random embeddings.
\newblock In \emph{IJCAI}, pages 1778--1784, 2013.

\bibitem[Weininger(1988)]{weininger1988smiles}
David Weininger.
\newblock {SMILES}, a chemical language and information system. 1. introduction
  to methodology and encoding rules.
\newblock \emph{Journal of chemical information and computer sciences},
  28\penalty0 (1):\penalty0 31--36, 1988.

\bibitem[Yang et~al.(2017)Yang, Hu, Salakhutdinov, and
  Berg-Kirkpatrick]{pmlr-v70-yang17d}
Zichao Yang, Zhiting Hu, Ruslan Salakhutdinov, and Taylor Berg-Kirkpatrick.
\newblock Improved variational autoencoders for text modeling using dilated
  convolutions.
\newblock In \emph{Proceedings of Machine Learning Research}, pages 3881--3890,
  2017.

\end{thebibliography}
\bibliographystyle{plainnat}

\end{document}